\documentclass[letterpaper, 10 pt, conference]{ieeeconf}  

\IEEEoverridecommandlockouts                              
\usepackage{graphicx}
\usepackage{color,soul}
\usepackage{mwe}
\usepackage{subcaption}
\usepackage{amsmath}
\usepackage{tabularx} 
\usepackage{multirow}

\newcolumntype{L}{>{\centering\arraybackslash}m{3cm}}
\usepackage{float}
\usepackage{graphicx} 
\title{\LARGE \bf
Pregrasp Object Material Classification by a Novel Gripper Design with Integrated Spectroscopy
}

\author{Nathaniel Hanson$^{1}$, Tarık Keleştemur$^{1}$, Deniz Erdoğmuş$^{1}$, Taşkın Padır$^{1}$
\thanks{This research is supported by the National Science Foundation under Award
Number 1928654.}
\thanks{$^{1}$Institute for Experiential Robotics, Northeastern University, Boston, Massachusetts, USA.}%
\thanks{$^{2}$Project repository: https://github.com/RIVeR-Lab/SpectroVision}%
\thanks{Nathaniel Hanson is the corresponding author. \newline{\tt\small hanson.n@northeastern.edu}}%
}

\begin{document}

\maketitle
\thispagestyle{empty}
\pagestyle{empty}

\begin{abstract}
Robots benefit from being able to classify objects they interact with or manipulate based on their material properties. This capability ensures fine manipulation of complex objects through proper grasp pose and force selection. Prior work has focused on haptic or visual processing to determine material type at grasp time. In this work, we introduce a novel parallel robot gripper design  and a method for collecting spectral readings and visual images from within the gripper finger. We train a nonlinear Support Vector Machine (SVM) that can classify the material of the object about to be grasped through recursive estimation, with increasing confidence as the distance from the finger tips to the object decreases. In order to validate the hardware design and classification method, we collect samples from 16 real and fake fruit varieties (composed of polystyrene/plastic) resulting in a dataset containing spectral curves, scene images, and high-resolution texture images as the objects are grasped, lifted, and released. Our modeling method demonstrates an accuracy of 96.4\% in classifying objects in a 32 class decision problem. This represents a performance improvement by 29.4\%  over the state of the art computer vision algorithms at distinguishing between visually similar materials. In contrast to prior work, our recursive estimation model accounts for increasing spectral signal strength and allows for decisions to be made as the gripper approaches an object.  We conclude that  spectroscopy is a promising sensing modality for enabling robots to not only classify grasped objects but also understand their underlying material composition.


\end{abstract}

\section{INTRODUCTION}

Humans have an excellent subconscious ability to recognize the materials around us throughout daily life. As we interact with our environment we leverage our vision and prior experiences to estimate the material properties of objects we come into contact with \cite{buckingham2009living}. As we reach out to pick up a coffee mug, we form an idea of what that object is likely to weigh and how its surface will feel. This knowledge directly informs the way we position our hands as we approach the mug, allowing us to plan for an appropriate grasp force, maintain contact with, and move the mug to wherever we desire. If the mug is composed of a fragile material like glass, our grasp strategy will likely change to support the object from the handle, or the thickest part. These steps constitute the Sense, Plan, Act (SPA) pipeline \cite{oddi2020integrating} for object manipulation. However for some materials, such as fruits or seafood, which are subject to irreversible deformations through contact, the sense step must be done with minimal object contact in a nondestructive manner.

Prior work on quantifying object material properties for robot applications contributed techniques based on human sensory capabilities, including touch and vision. Specifically in prior work involving direct contact with objects (presented in detail in Section \ref{sec:prior}), much attention has been given to visual sensing or texture estimation via an elastomeric sensor.

\begin{figure}[t]
  \centering
  \includegraphics[width=\linewidth]{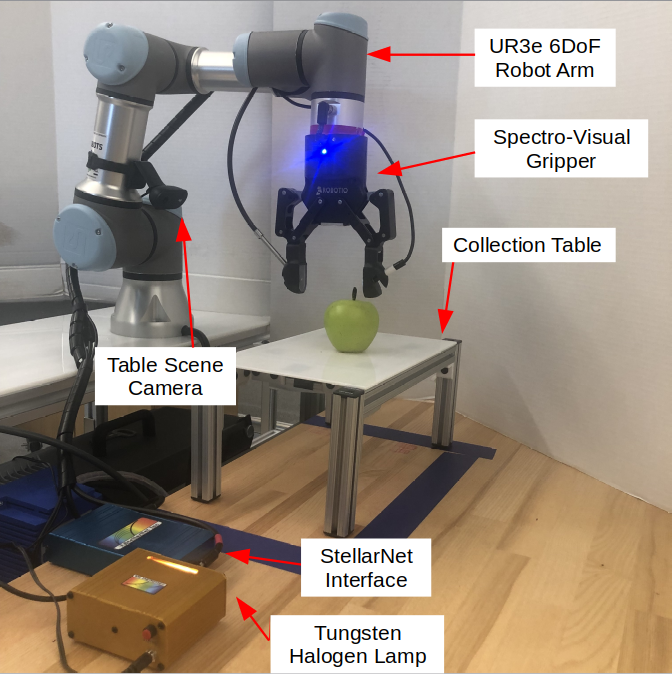}
  \caption{Data collection setup using integrated spectral and visual perception.} 
  \label{cross_collection_setup}
  \vspace{-2.5em}
\end{figure}
In this research, we present the design of a two fingered gripper$^{2}$ capable of collecting images and spectral data in real-time as the robot approaches an object to be grasped. This allows the robot to collect data and infer object material type before contact is made. Our research relies on the corpus of prior work in Near-Infrared (NIR) spectroscopy \cite{pasquini2018near, bec2019advances}, which has been demonstrated as a reliable means to estimate the material properties and composition of an object through reflected light in the ultraviolet ($\lambda \approx 300\ \text{nanometers}$) to the near-infrared ($\lambda \approx 1200\ \text{nanometers} $) spectra. Instead of representing light with three discrete values as is commonly done with red, blue, green (RBG) image channels, the integrated spectrometer represents light as a continuous curve, sampled at 2048 discrete values -- allowing for the observation and quantification of changes in the spectral response of an object. We classify incoming spectral readings with an SVM and leverage recursive estimation to increase confidence in the class label as more data is acquired.\par
The contributions of this paper are as follows.
\begin{itemize}
    \item We introduce a novel design for a two fingered gripper with an integrated NIR spectrometer and endoscopic camera to collect VNIR spectral readings and macro surface images from grasped items.
    \item We develop a method based on a nonlinear Support Vector Machine (SVM) to achieve material inference between visually similar items and continually update class estimates with a discrete Bayes filter.
\end{itemize}

The paper is organized as follows. Section~II provides an overview of the prior work in robot material sensing through visual and haptic approaches. To provide rationale for the spectro-visual gripper, we offer a brief discussion on the capabilities and underlying science of visible-NIR spectroscopy in Section~III. Section~IV is dedicated to presenting the details of the gripper design. SVM-based classification method is presented in Section~V. Section~VI discusses the validation of the model and experimental results on the classification of organic and inorganic materials with machine learning. Finally, Section~VII discusses the results of our research and future opportunities for spectral sensing.

\section{RELATED WORK}
\label{sec:prior}

\subsection{Material Identification}
Sensing modalities for classifying materials falls into two categories: contact and contact-free sensing. Computer vision constitutes the majority of work in contact-free material sensing. \cite{schwartz2019recognizing} hypothesized that the subjective labels of material appearance, such as  fuzzy, shiny, smooth, rough, from human labeled regions could be used in a Convolutional Neural Network (CNN) to segment images intro broad material regions with reasonable success. Similarly, others have utilized a CNN \cite{bell2015material} to classify patches by material type as an input to a fully connected random field to contextualize the material type to the surrounding environment. These image based techniques require tedious and time-consuming hand segmentation and essential ground truth information. For example, plastics, metals, and ceramics can all look similar when painted and illuminated with specular reflection \cite{fang2019toward}.\par
In contact sensing, surface textures and shapes are exploited to determine the material type. For instance, a 3D printed finger was devised for use by humanoid robots to collect accelerometer data as the finger was moved across the surface of various material types \cite{sinapov2011vibrotactile}. Using a hierarchical model leveraging the frequency-domain signal from accelerometer, an SVM achieves an 80.0\% accuracy in recognizing surface textures. This  framework has also been shown to be extensible to new material types and hence, it can categorize various materials based on similarities in texture and friction coefficients. While a novel approach to material classification, the exploratory behaviors are slow and can take upwards of 30 seconds to complete before the classification can begin and require direct contact. Related work in haptic sensing \cite{takamuku2007haptic}, and its fusion with visual data \cite{strese2016multimodal} has demonstrated promising results on a variety of everyday objects.\par
Elastomeric sensors measure surface topography and contact forces through deformation of a contact gel imaged by a small, high-resolution camera. The GelSight sensor is commonly used to acquire high-resolution texture models for fabrics and other coarse surfaces \cite{li2013sensing}, \cite{wang2016robot}. Force feedback from the gel deformation and CNNs have been used to learn manipulation models for a variety of material types \cite{abad2020visuotactile}. While these sensors provide rich surface information, it is computationally heavier to do material classification due to high-dimensional data. In contrast, spectral data contains and order of magnitude less data but still achieves high classification accuracy.
\subsection{Spectroscopy}
Previous research in spectral material identification is centered on commercial, handheld spectrometers fitted to a humanoid robot to scan household objects and differentiate between metal, plastic, wood, paper, and fabric \cite{erickson2019classification}. With a neural network architecture this model resulted in a classification accuracy of 79.1\% and was shown to generalize well to previously unseen objects \cite{erickson2020multimodal}. The approach was further extended to determine object penetrability for tool construction from multiple objects \cite{nair2019autonomous}. Agriculture researchers also developed a specialized gripper with a spectral probe and force feedback to determine the quality and ripeness of mango fruits through simultaneous spectral sampling and force feedback\cite{cortes2017integration}. Their combination of accelerometer and spectral data resulted in the creation of a regression model which could accurately predict the ripeness of a mango. However, their gripper was specifically designed to fit mangoes and did not generalize to other applications.\par

Our work is differentiated from the spectroscopy solution presented in \cite{erickson2019classification} and \cite{erickson2020multimodal} since these works regard spectral readings as a static measurement to be acquired at a fixed position from the object and over a single time step. Our work integrates spectral probes directly into the gripper finger, providing an avenue to both sense and grasp in one continuous motion, instead of scanning, and grasping separately. This combination of perception and manipulation reduces the total time needed to plan interactions with items. Moreover, we leverage the fast acquisition time of our spectrometer to continually improve our material classification as the gripper fingers close around the object. In contrast to existing work in haptic perception, our methodology does not require any form of contact and allows material perception on objects which might deform under contact.
\section{BACKGROUND ON SPECTROSCOPY}
NIR spectroscopy is a technique which involves measuring the electromagnetic radiation emitted by an object when illuminated in the range of 750 to 2500 nm. Chemically, NIR spectroscopy occurs due to the anharmoic vibrational modes caused by the presence of heavy elements such as Carbon, Nitrogen, Oxygen bonding to a Hydrogen atom or bonds between these molecule types \cite{pasquini2018near}. As electromagnetic energy is imparted on a target substance, the bonds react to the energy in the NIR range and causes overtones in the bond vibrations. All materials exhibit some form of NIR reflectance and absorption, similar to how materials reflect red, blue, and green light in varying proportions, which our retinas and their cone cells process into color rich environmental information. For both organic and inorganic materials composed of these aforementioned bonds, particularly in long chains or rings, these substances exhibit markedly unique responses in the amount of non-visible NIR radiation they emit. By measuring the response profiles of material samples, future acquisitions can be compared to this reference spectral library to determine material class \cite{kokaly2017usgs}.\par
In analytical spectroscopy a full spectrum light source, marked by even electromagnetic emittance across the spectrum, is used to illuminate a surface. A diffraction grating, akin to a prism, separates the light into component wavelength ranges which measures incident photons using either a Silicon or Indium Gallium Arsenide (InGaAs) detector array - depending on the wavelength being measured. NIR spectroscopy has found widespread adoption in the laboratory and research realms. In agriculture, processing quality checks regularly rely on NIR to estimate sugar, fat, and overall quality of foodstuffs \cite{kawasaki2008near},\cite{shah2020towards}, \cite{guermazi2014investigation}.
\section{METHODS \& MATERIALS}
The gripper design process has been driven by a set of design specifications which include: (1) Handle objects of size 0-8.5 cm in diameter; (2) Collect continual spectral data at a rate of 10 Hz in the wavelength range 400 - 1100 nm; (3) Acquire texture images from a range from 1 - 8.5cm.
\subsection{Spectrometer Specifications}
We utilize a StellarNet BLUE-Wave miniature spectrometer: a compact, comparatively less expensive device has been shown to perform well against more expensive laboratory devices and is well utilized in the literature \cite{pourdarbani2020non}, \cite{jenkins2012molecularly}. Our system design utilizes a seven strand fiber optic cable connecting the device to a cylindrical probe and collects readings from 350 - 1150 nm. A single central fiber is used to measure reflectance and the other six fibers connect to a Quartz Tungsten Halogen (QTH) lamp and provide the requisite full spectrum illumination. This QTH light source is necessary to ensure uniform operation of the spectral sensor across a variety of lighting environments. The device is calibrated with a white reference standard to account for a perfect pure white light reflection, and a probe covering to establish a dark signal baseline. The Signal to Noise ratio (SNR) is 1000:1, and the signal quality can be further improved by increasing integration time (time to acquire reflectance information) and signal sample averaging. The device was sampled at a rate of 10Hz and the data is transmitted to a desktop computer. Our fiber optic cable has a bend radius of 19.8 cm, which we are careful to avoid exceeding as it could cause internal damage to the cable, or result in signal attenuation from internal refraction.
\subsection{Gripper Design}
\begin{figure}[t]
  \vspace{0.5em}
  \centering
  \includegraphics[width=\linewidth]{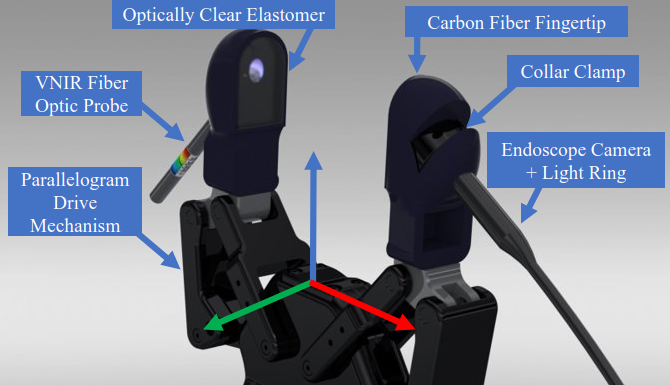}
  \caption{Gripper with 3D printed finger pads, an NIR spectrometer probe and endoscope attachments rendered with AutoCAD.} 
  \label{gripper_design}
  \vspace{-1.0em}
\end{figure}

Using a Robotiq 2F-85 parallel plate gripper, we designed custom finger pads to extend the base mechanism's capabilities to collect integrated spectral grasping setup. Each finger measures (64 mm x 30 mm x 27 mm) with a contact pad (35 mm x 22 mm x 3 mm), with 23.1 square centimeters of intended contact space. The two finger pads, pictured in Fig \ref{gripper_design}, are 3D printed with continuous carbon fiber reinforcement to minimize flexing at the mounting point of the finger to the parallelogram drive mechanism. The fingers have angled circular slots to allow for the insertion and adjustments of probes with a friction fit collar clamp. The spectrometer probe was positioned so 1 cm separation remains when the robot grasps an object. The other finger is fitted with an endoscope camera providing high resolution images from the perspective of the finger pad. It also contains a brightness-adjustable Light Emitting Diode (LED) light ring to provide proper illumination. Both devices are inserted at a $45^{\circ}$ angle relative to grasping surface to ensure good coverage of grasped materials while minimizing the width of the gripper.\par
\begin{figure}[t]
  \centering
  \includegraphics[width=\linewidth]{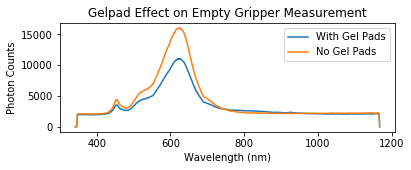}
  \caption{Spectral reflectance curves collected with and without the presence of finger gelpad showing uniformly diminished reflectance due to minor subsurface scattering}
  \label{gel_effect}
  \vspace{-1.5em}
\end{figure}

To ensure a gentler grasp and protect the sensors, the fingers contain inserts inspired by the DIGIT tactile sensor made of Solaris silicon gel \cite{lambeta2020digit}. This material was selected for its wide operational temperature range of $-100^{\circ}$ and $204^{\circ}$ C and resilience when grasping rough textured objects. Moreover, its rated tensile strength exceeds the force capabilities of the Robotiq gripper. Notably, the gel is also optically clear in the operating range of the BLUEWAVE spectrometer and verified by a hyperspectral camera (Headwall Photonics) to not significantly impede the transmission of any visible or NIR light as shown in Fig. \ref{gel_effect}.
\subsection{Data Collection}
\begin{figure*}[t]
    \centering
    \vspace{0.5em}
    \includegraphics[width=\textwidth]{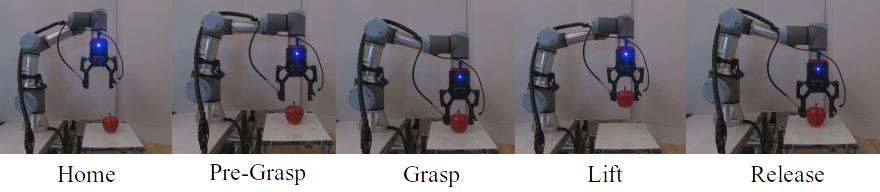}
    \caption{Grasping procedure for collecting spectral and image data with a robotic manipulator through defined series of waypoints and end effector positions.}
    \vspace{-1.65em}
    \label{collection_procedure}
\end{figure*}
Our experimental data was acquired using a UR3e collaborative robot arm controlled by the Robot Operating System (ROS) \cite{quigley2009ros}. The robot has four predefined positions defined in the Cartesian space: HOME, PRE-GRASP, GRASP, and RELEASE. These positions are shown in the sample data collection in Fig. \ref{collection_procedure}. Each position consisted of an $[x,y,z]$ coordinate and quaternion representing the end effector pose. To transition between positions, we use an inverse kinematics solver \cite{wampler1986manipulator} that respects the joint limits and 3D mesh of the robot model. Objects were placed at the center of a white table, surrounded by a white background to minimize ambient reflections into the spectrometer. An RGB camera was mounted to the robot to capture mages of the full table scene. Fig. \ref{cross_collection_setup} shows the complete collection environment. The gripping force threshold was set at 2.5\% (5.9 N) of the maximum possible force (235 N) to avoid damaging the fruits. Objects were lifted to induce small variations into the spectral signal by flexing the fiber optic cable.\par
For the initial demonstration of the system capabilities, we designed a visually challenging experiment for the spectro-visual gripper. We selected 16 varieties of fresh fruit \textit{\{kiwi, banana, green grape, lemon, mango, avocado, plum, orange, red apple, pear, strawberry, red grape, black grape, green apple, peach\}}. For each real item of fruit, we acquired a corresponding fake item. These items are designed to look visually indistinguishable  from their real counterparts and share similar surface textures.\par
For each sample acquisition, illustrated in Fig. \ref{collection_procedure}, the object is placed on the table in front of the robot. The robot takes a single image using the full scene camera. The robot then transitions from HOME to PRE-GRASP where it is directly above the object. The arm then lowers to place its gripper fingers on either side of the object to the GRASP position. The endoscope is triggered and takes a texture picture of the object to be grasped. As the gripper fingers close, the spectral data acquisition commences. Once the fingers detect force feedback exceeding the threshold, the arm lifts the object back to PRE-GRASP position, holds for one second and then lowers to release the item, ending spectral data collection. This procedure is repeated 5 times for each object, with the object rotated $20^{\circ}$ between collections.\par
\begin{figure}[b]
  \vspace{-1.5em}
  \begin{subfigure}{0.10\textwidth}
    \includegraphics[width=\linewidth]{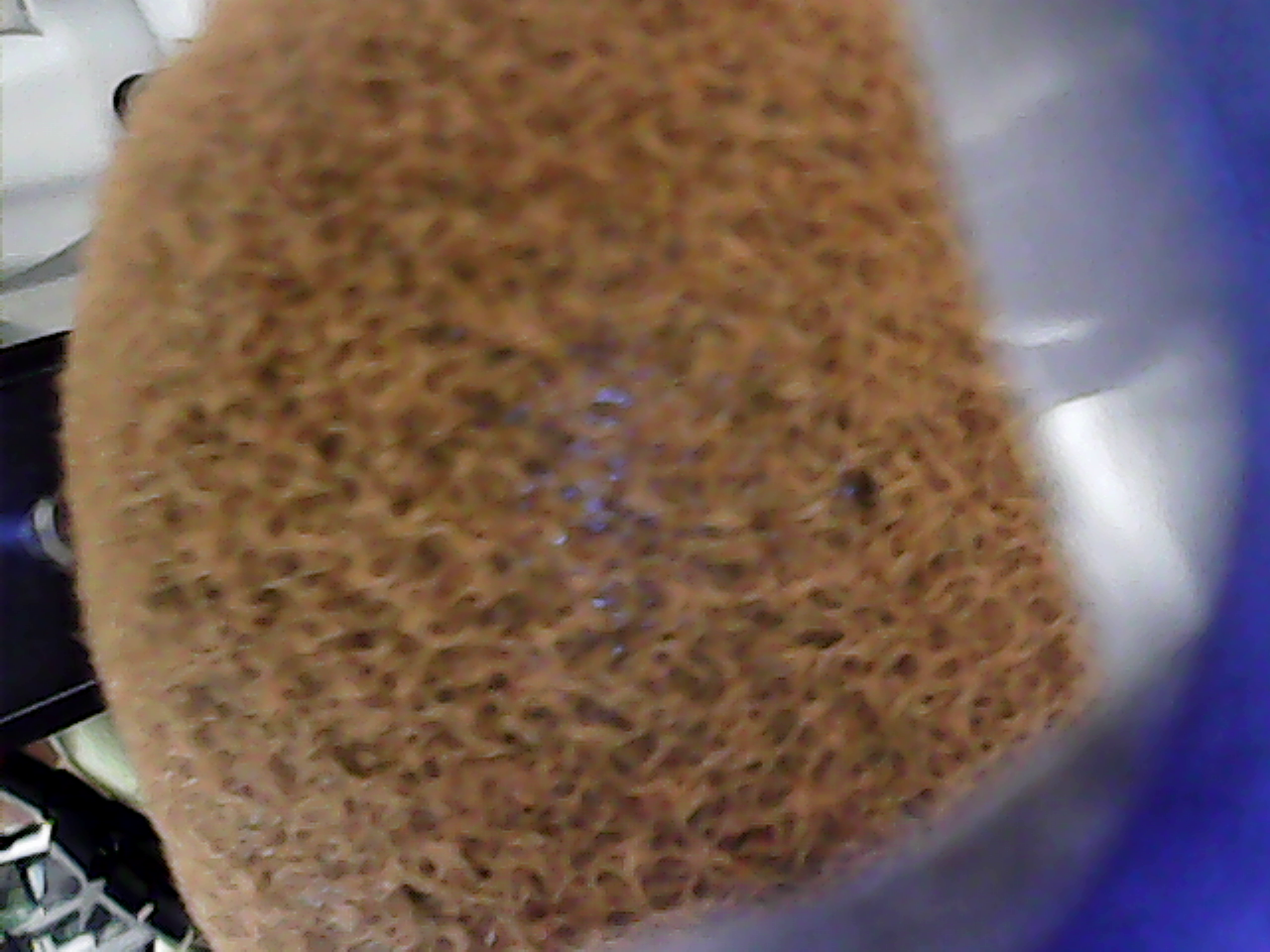}
    \caption{Fake kiwi texture} \label{fake_endo_1}
  \end{subfigure}%
  \hspace*{\fill}   
  \begin{subfigure}{0.10\textwidth}
    \includegraphics[width=\linewidth]{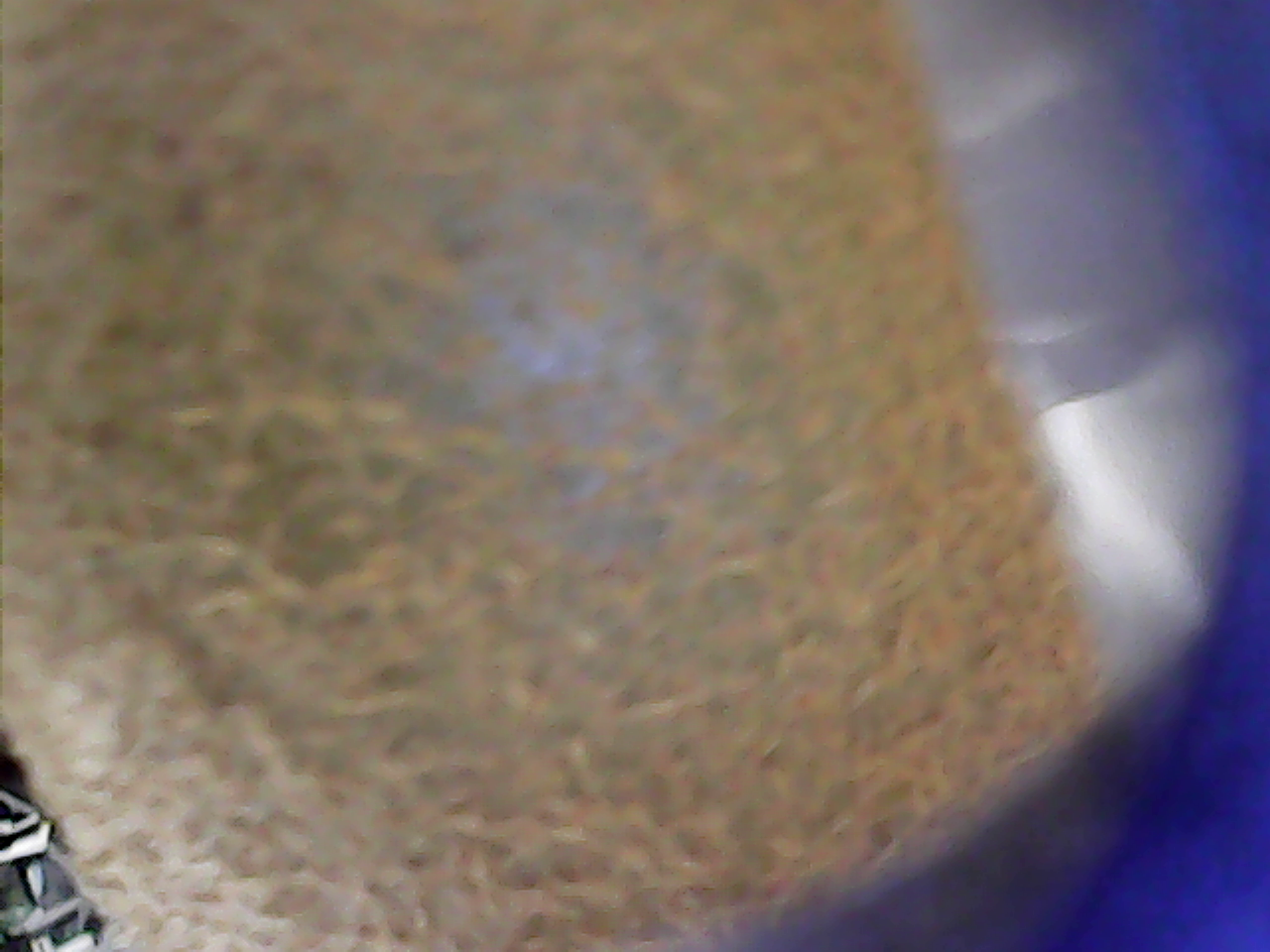}
    \caption{Real kiwi texture} \label{real_endo_1}
  \end{subfigure}
  \hspace*{\fill}   
  \begin{subfigure}{0.10\textwidth}
    \includegraphics[width=\linewidth]{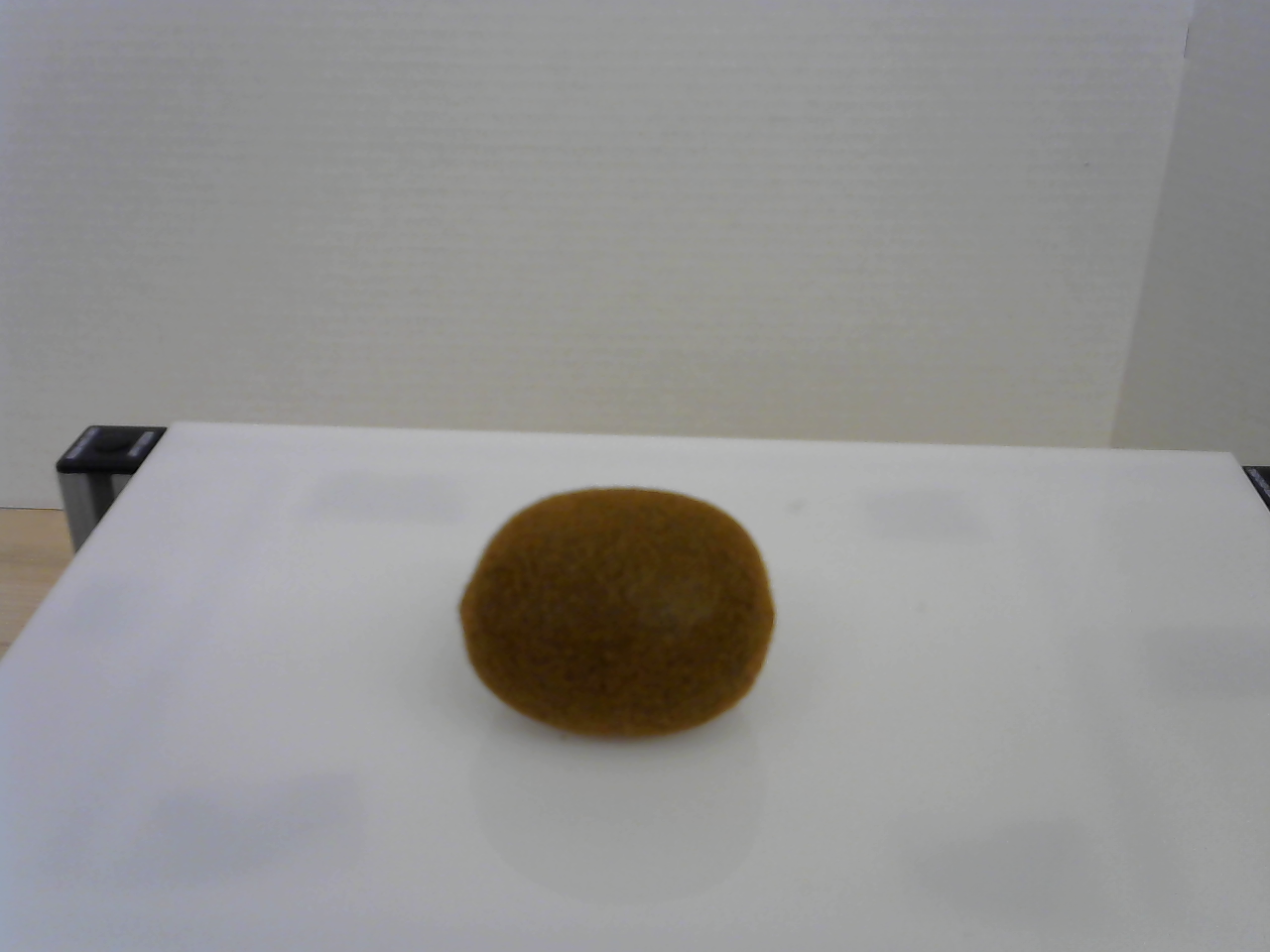}
    \caption{Full fake fruit} \label{fake_scene_1}
  \end{subfigure}
  \hspace*{\fill}   
  \begin{subfigure}{0.10\textwidth}
    \includegraphics[width=\linewidth]{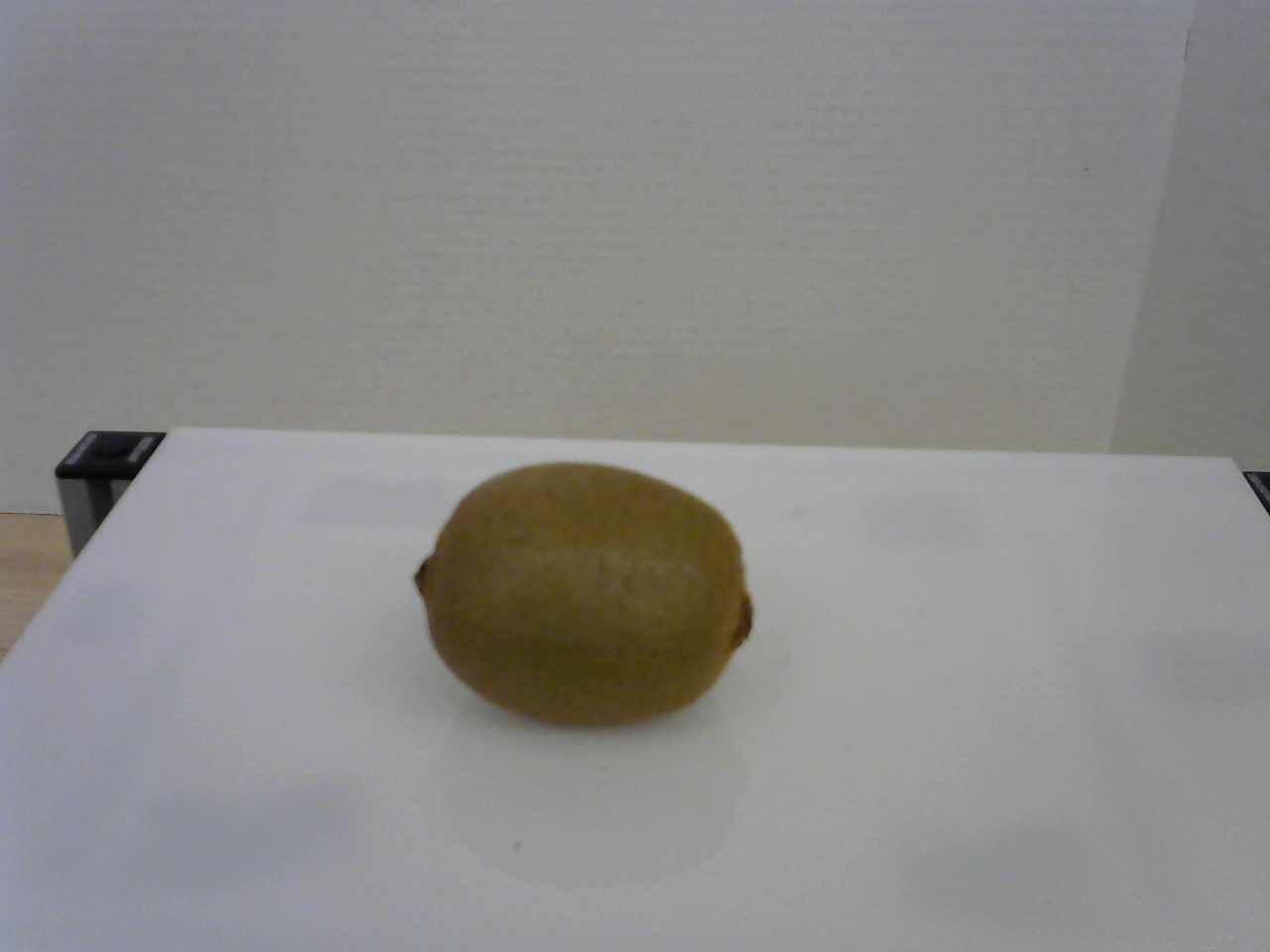}
    \caption{Full real kiwi} \label{real_scene_1}
  \end{subfigure}%
\caption{Images collected by system during grasping. Note the similar size, shape, color, and textures of the two fruits.} \label{collection_images}
\end{figure}

In Fig. \ref{collection_images} both the fake and real kiwi posses similar light brown surface hair. The fake items are composed of a variety of plastics and other inorganic substances contrasted with the organic structure of the real fruits. The real fruits are composed primarily of water and carbohydrates, whose molecular structure differs significantly from plastics and polystyrene counterparts \cite{zhang2019hyperspectral}. As a result, we create a 32 class problem containing each of the real and fake varieties.
\section{MODELING}
Our main goal is to classify objects using spectroscopy and show that this technique outperforms vision-based methods in a real world setting with visually similar items. To this end, we first deploy a state-of-the art visual encoder and train it on the dataset we collected within the scope of this research. Later, we discuss our method for using spectral information that can achieve better results with simpler models. In general, the problem can be formulated as a multi-class classification:
\begin{align}
\hat{x} = \textit{argmax} (p_{\theta}(x|z))
\end{align}
where $\hat{x}$ is the predicted class, $\theta$ is the model parameters, and $z$ is the observation (e.g. visual image or reading from spectrometer). In the case of real-world experiments, we use recursive estimation to update the belief over the material classes. This way, we can predict the true class using the history of observations as opposed to doing single time predictions during the whole grasping procedure.
\subsection{Vision-based Classification}
As a classification baseline, we leverage a CNN to perform whole image classification with the scenes collected from the full-scene camera and endoscope. We capture full scene images to provide an analogous visual to how humans would classify a fruit: by using the whole object and its shape, size, and color. The endoscope provides a macro look at the fruits' textures and surface coloration that might not be visible from the scene camera's location. Within the same table environment we collected 100 images from orbital rotations around the fruit centered on the collection table. We also rotated the fruit within the open gripper and captured 100 macro texture images of each fruit with the endoscope.\par
To increase the robustness of our visual model to noise, we augment our existing dataset with the following transformations: random crops, random affine transformations, random horizontal flips, random color equalization, and random sharpeni\cite{shorten2019survey}. We leverage a pre-trained ResNet-50 architecture \cite{he2016deep} on the ImageNet \cite{deng2009imagenet} with the last fully connected layer altered to classify the 32 classes in our decision problem. Our model is trained in PyTorch \cite{paszke2019pytorch} with the Adam optimizer \cite{kingma2014adam} with parameters of $\beta_{1} = 0.9$, $\beta_{2} = 0.999$ and a learning rate of 0.0001. The model is trained for 50 epochs with a decaying learning rate every 15 epochs. This same architecture was used to classify both endoscope and full-scene images.
\subsection{Spectroscopy}
The continuous spectral curve for each sample acquisition consisted of 2048 discrete samples, as limited by the diffraction gratings on the spectrometer detector. For each collection iteration, $\approx65$ spectral samples were collected from the start of the PRE-GRASP phase to the completion of the RELEASE phase. Scan averaging helped minimize the noise in the data in addition to a $5^{th}$ degree Savitzky-Golay filter. After removing a few erroneous data entries, the collected data was stacked into a matrix with dimensionality $10443\ \times \ 2048$. Knowing the high dimensionality of data would slow inference time for incoming signals, we reduce the dimensionality of the data using Non-Negative Matrix Factorization (NMF)~\cite{dhillon2005generalized} to learn a series of basis vectors $H$. This matrix yields a transformation of the original dataset into an $n$-dimensional array, whose length is a selected hyperparameter. We selected NMF over Principal Component Analysis (PCA) and Singular Value Decomposition (SVD) as our matrix decomposition technique since it is guaranteed to yield strictly positive factors, which is desirable as spectral signals are also completely positive.\par
We train four different model types: logistic, linear kernel Support Vector Machine (SVM), Radial Basis Function (RBF) nonlinear SVM, and Multi-Layer Perceptron (MLP). We utilized 5-fold cross validation with grid-search hyperparamter selection to identify appropriate regularization and kernel constraints for each model while using 80\% of the data for training while maintaining class balance. After selecting the hyperparameters which maximize the average performance across k-fold training on the validation data, we train a final iteration of models on the entire training data and evaluated them using the held-out $20\%$ of data.
\begin{figure}[t]
  \centering
  \includegraphics[width=\linewidth]{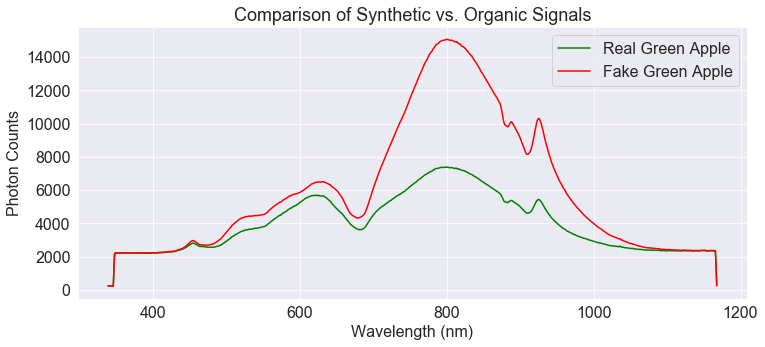}
  \caption{Real vs. synthetic spectral profiles for green apples types. These curves demonstrate the mean profile of all the spectral signals in each class.}
  \label{real_vs_synthetic}
  \vspace{-1.5em}
\end{figure}
\subsection{Recursive Estimation}
As the spectral-finger nears the target object, the spectral reading changes until it reaches a stable position in the GRASP position. When the gripper is empty, the signature is identified as the reflectance of overhead LED lighting entering the sensor. This signature is consistent despite variations in the end effector pose. When the robot begins a grasping procedure, we assume a uniform prior distribution across all classes. We hold the Markov assumption, and formulate a recursive estimation model for incoming spectral signals. While we can do point estimate at every single timestep, recursive estimation allows us to use the whole history observations and overcome any possible noise in a particular timestep. We use a discrete Bayes filter (or Histogram filter)~\cite{thrun2002probabilistic} to maintain a belief over the possible materials. Let $x_t$ be the material type, $z_t$ be the measurements, and $bel(x_t)$ the belief at time step $t$ which is represented as a vector. Then, we do observation updates as follows:
\begin{align}
    bel(x_t) = \eta p(z_t|x_t)bel(x_{t-1})
\end{align}
where $\eta$ is the normalization factor and $p(z_t|x_t)$ is the observation likelihood probability function. Note that the Bayes filters usually have a prediction step consisting of the the motion model but we skip this step as the state of our problem (material type) does not change during the prediction. Our observation model is modeled using machine learning methods that are described above.

When the gripper first approaches the object, the ambient lighting and the color in the visible spectrum are the first parts of the spectral curve exhibit notable response. At this point, the model might initially misclassify two similarly colored objects. As the gripper closes, the amplitude of the NIR segment becomes the curve's most prominent feature. This observation is confirmed by the calibration documents provided by the optical gratings manufacturer, as the silicon detector is less effective in the upper range of the NIR spectrum than the visible light range. However, proximity to the sensor appears to counteract this intrinsic sensitivity. 

This update rule enables the robot gripper to gain increasing confidence in the material before contact. Once a threshold $\kappa$ confidence level is  exceeded or $n$ samples have been taken, the object is assigned a label. 
\begin{figure}[t]
  \centering
  \includegraphics[width=\linewidth]{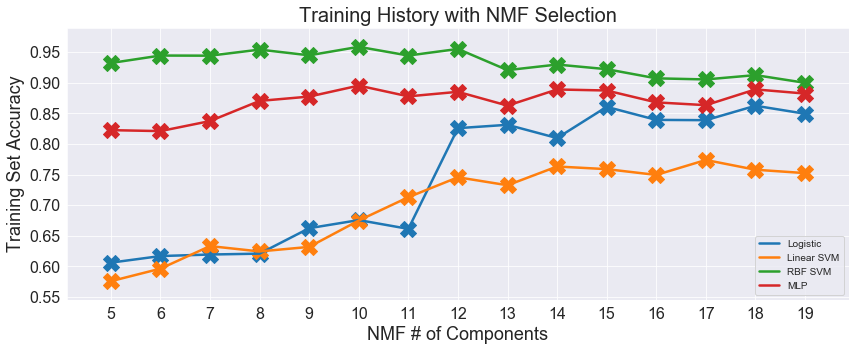}
  \caption{Training history for average model performance for each number of NMF components} 
  \label{train_hist}
  \vspace{-1.5em}
\end{figure}
Fig. \ref{real_vs_synthetic} shows the spectral data acquired for two visually indistinguishable green apples. The nearly identical spectral curves in $\lambda \rightarrow (350, 675)\ \text{nm}$ show similar visible light patterns, indicating their visible color is nearly identical. However, the NIR range $(>700\ \text{nm})$ shows a large intensity spike not perceptible to the human eye. This same NIR region could be compared with existing databases of spectral signals for more precise material composition analysis and similarity analysis to other cataloged materials \cite{kokaly2017usgs}.
\section{RESULTS}
The overall performance metrics are reported in Table \ref{res_table}. The RBF SVM maintains strong performance even when working with limited components from the NMF. Similarly the MLP also performs well with approximately the same number of components. This observation reinforces that the spectral classification problem is a nonlinear model, but is ultimately well explained through the introduction of nonlinearities in MLP layer connections and the kernel transformation by the RBF.\par
The endoscope classifier achieved an accuracy of 0.67 on 1920 test images. The full scene classifier performed worse with an accuracy of 0.59 over 2669 test images. Despite the success of ResNet in general image classification problems where there are significant inter-class differences, this task highlights the limitations of solely using visual sensing.\par
In Fig. \ref{train_hist} we observe how NMF performs as a dimensionality reduction technique on spectrometer data. For the RBF SVM, 10 components is maximally efficient and reduces our input data by $99.5\%$. Although the number of input components is fewer than the number of classes, the small input size indicates there is sufficient variations in intensity between the components to have separability in the classes. The small input size also has noticeable effects on the inference time to process new samples. Our trained ResNet model is 93.3 megabytes, compared with 45.8 kilobytes for the SVM, and is approximately two orders of magnitude slower to process incoming data than the RBF SVM. Table \ref{time_table} shows the mean and standard deviations of the model inference times evaluated over the test datasets. Moreover, the SVM does not require a GPU to run at realtime processing speeds, thus decreasing the computing cost.\par
\begin{table}[t]
    \vspace{0.5em}
    \caption{Model results for spectral and image models.}
    \label{res_table}
    \centering
    \begin{tabular}{||c|c|c|c|c||}
     \hline
     Model & NMF & Train Acc & Test Acc & Test F-1\\ [0.5ex] 
     \hline\hline
     \multirow{2}{*}{} & \multicolumn{4}{c||}{Spectral}\\
     \hline
     Logistic & 18 & 86.23 & 82.72 & 80.02\\ 
     \hline
     Linear SVM & 17 & 77.38 & 89.13 & 90.06\\
     \hline
     RBF SVM & 10 & 95.85 & \textbf{96.41} & \textbf{95.44}\\
     \hline
     MLP & 10 & 89.50 & 91.48 & 88.35\\
     \hline
     \multirow{2}{*}{} & \multicolumn{4}{c||}{Vision}\\
     \hline
     Full-Scene & - & 81.00 & 59.00 & 53.82\\
     \hline
     Endoscope & - & 89.00 & 67.00 & 65.92\\
     \hline
     \end{tabular}
     \vspace{-1.75em}
\end{table}

Specifically with respect to the hyperparameter ranges for the MLP, we selected for both the number of layers and the number of their constituent neurons. There are many deeper networks with general success in machine learning; however, these models exhibit the same size and complexity problems as ResNet. Similarly, there are a wide variety of kernel methods for nonlinear SVMs which might provide additional performance improvements.

\begin{table}[t]
    \vspace{0.5em}
    \caption{Model inference times for visual and spectral models on Intel i7 processor with GTX 1070 GPU.}
    \label{time_table}
    \centering
    \begin{tabular}{||c || c | c||}
     \hline
     Model & $\mu_{\text{inference}}$ (s) & $\sigma_{\text{inference}}$ (s)\\ [0.5ex] 
     \hline\hline
     RBF SVM & 0.00019 & 0.00036\\
     \hline
     ResNet (GPU) & 0.01928 & 0.00159\\
     \hline
     ResNet (CPU) & 0.70733 & 0.06923\\
     \hline
    \end{tabular}
    \vspace{-2.0em}
\end{table}
Our results are comparable to prior spectral-based model for the classification of household materials \cite{erickson2019classification}, although our recursive model adds additional estimation capabilities and functions on a greater class set size (32 versus 5). In comparison to accuracy of the visual methods using ResNet, the SVM with spectral data exceed the best performing endoscope model by 29.4\%. Each class in the decision problem is classified with an accuracy  $\geq88\%$; moreover, for the misclassified examples, the error is associated with the distance from the object as discussed in Section~V. Utilizing recursive estimation, were were able to further increase the accuracy to 100\% using only three consecutive spectral samples for each object type.
\section{CONCLUSIONS}
In this work, we presented a novel gripper design to incorporate spectral sensing in a compact and integrated form. We demonstrated this system as a means to acquire a meaningful chemical profile via spectroscopy. With inspiration from elastomeric sensing, we designed a robust gripper capable of material sensing beyond the capabilities of humans without the need to contact the object directly. We evaluated our design on an assortment of 16 pairs of real and fake fruits, demonstrating the gripper was able to acquire visible-NIR spectral readings leading to successful item classification without the need for direct contact with the object. Compared with prior work in tactile sensing, our design and methodology allows for contact-free material determination. With recursive estimation, our model provides increased confidence in material class and can account for sensor noise while converging with enough time for the grasp to be aborted or changed.\par
We envision miniaturizing this technology to fit entirely within the fingerpads on a custom Printed Circuit Board (PCB) in future iterations of this gripper design. The sensor we utilized only provides a small look into the NIR, as the consensus for NIR range extends out to 2500 nm. Expanding the sensing abilities spectrometer by switching to an InGaAs detector will allow for easier differentiation between broad material categories, visibly transparent materials, and beneath surface coatings \cite{petersen2021non}.\par
As the gripper already contains an endoscope there is potential for multi-modal sensing with spectral data with either high-resolution texture imagery or other forms of haptic feedback. 
Moreover, there are well defined databases of spectral material signatures with high degrees of specificity. Incorporating those databases into a trained model for grasped material identification will be useful for robots manipulating unknown objects or objects composed of heterogeneous materials.\par
\section*{ACKNOWLEDGMENTS}

The authors are grateful to Iris Wang and Stephanie Hanson for their help in the data collection, Hillel Hochsztein for his CAD and machining knowledge, and Charles DiMarzio for his expertise in spectroscopy.




\bibliographystyle{IEEEtran} 
\bibliography{references}
\end{document}